\documentclass[10pt,twocolumn,letterpaper]{article}

\usepackage{iccv,times}
\def\iccv{for iccv submission}

\usepackage{color,xcolor}
\usepackage{epsfig}
\usepackage{graphicx}

\usepackage{tikz}
\usetikzlibrary{arrows}

\usepackage{chngcntr}

\usepackage{tabularx}
\usepackage{adjustbox}
\usepackage{array}
\usepackage{booktabs}
\usepackage{colortbl}
\usepackage{float,wrapfig}
\usepackage{hhline}
\usepackage{multirow}
\usepackage{nidanfloat} 

\usepackage{amsmath,amsfonts,amsthm,amssymb}
\usepackage{bm}
\usepackage{nicefrac}
\usepackage{microtype}
\usepackage{dsfont}
\usepackage{changepage}
\usepackage{extramarks}
\usepackage{fancyhdr}
\usepackage{lastpage}
\usepackage{setspace}
\usepackage{soul}
\usepackage{xspace}

\usepackage[pagebackref=true,breaklinks=true,colorlinks,citecolor=black,urlcolor=blue,bookmarks=false]{hyperref}

\usepackage{algorithm, algorithmic}
\usepackage{paralist}
\usepackage{enumitem}

\usepackage[pagebackref=true,breaklinks=true,colorlinks,citecolor=black,urlcolor=blue,bookmarks=false]{hyperref}

\usepackage{algorithm, algorithmic}
\usepackage{paralist}
\usepackage{enumitem}
\usepackage{todonotes} 

\newcommand{\citep}[1]{\cite{#1}}
\newcommand{\citet}[1]{\cite{#1}}

\usepackage{amsmath,amsfonts,bm}

\def\rvr{{\mathbf{r}}}

\def\rvx{{\mathbf{x}}}

\def\rvz{{\mathbf{z}}}

\DeclareMathAlphabet{\mathsfit}{\encodingdefault}{\sfdefault}{m}{sl}
\SetMathAlphabet{\mathsfit}{bold}{\encodingdefault}{\sfdefault}{bx}{n}

\def\gL{{\mathcal{L}}}

\usepackage{amsmath,amsfonts,bm}

\newcommand{\fid}{Fr\'echet Inception Distance\xspace}
\newcommand{\wsd}{Wasserstein Sliced Distance\xspace}
\newcommand{\gissfull}{Generated Image Segmentation Statistics\xspace}
\newcommand{\fssshort}{FSD\xspace}
\newcommand{\fssfull}{Fr\'echet Segmentation Distance\xspace}

\newcommand{\final}{later\xspace}
\newcommand{\dist}{\ell}
\newcommand{\disti}{\dist}

\newcommand{\reffig}[1]{Figure~\ref{fig:#1}}
\newcommand{\refsec}[1]{Section~\ref{sec:#1}}

\newcommand{\reftbl}[1]{Table~\ref{tbl:#1}}

\newcommand{\lblfig}[1]{\label{fig:#1}}
\newcommand{\lblsec}[1]{\label{sec:#1}}

\newcommand{\lbltbl}[1]{\label{tbl:#1}}

\newcommand{\ignorethis}[1]{}

\newcommand{\myparagraph}[1]{\vspace{-5pt}\paragraph{#1}}

\def\eqref#1{equation~\ref{#1}}

\def\1{\bm{1}}

\def\rvr{{\mathbf{r}}}

\def\rvx{{\mathbf{x}}}

\def\rvz{{\mathbf{z}}}

\DeclareMathAlphabet{\mathsfit}{\encodingdefault}{\sfdefault}{m}{sl}
\SetMathAlphabet{\mathsfit}{bold}{\encodingdefault}{\sfdefault}{bx}{n}

\def\gL{{\mathcal{L}}}

\newcommand{\E}{\mathbb{E}}

\newcommand{\range}{\mathrm{range}}

\DeclareMathOperator*{\argmin}{arg\,min}

\DeclareMathOperator{\Tr}{Tr}

\newcolumntype{L}[1]{>{\raggedright\let\newline\\\arraybackslash\hspace{0pt}}m{#1}}
\newcolumntype{C}[1]{>{\centering\let\newline\\\arraybackslash\hspace{0pt}}m{#1}}
\newcolumntype{R}[1]{>{\raggedleft\let\newline\\\arraybackslash\hspace{0pt}}m{#1}}

\newcommand{\ignore}[1]{}

\makeatletter
\DeclareRobustCommand\onedot{\futurelet\@let@token\@onedot}
\def\@onedot{\ifx\@let@token.\else.\null\fi\xspace}

\makeatother

\definecolor{MyDarkBlue}{rgb}{0,0.08,1}
\definecolor{MyDarkGreen}{rgb}{0.02,0.6,0.02}
\definecolor{MyDarkRed}{rgb}{0.8,0.02,0.02}
\definecolor{MyDarkOrange}{rgb}{0.40,0.2,0.02}
\definecolor{MyPurple}{RGB}{111,0,255}
\definecolor{MyRed}{rgb}{1.0,0.0,0.0}
\definecolor{MyGold}{rgb}{0.75,0.6,0.12}
\definecolor{MyDarkgray}{rgb}{0.66, 0.66, 0.66}

\iccvfinalcopy 

\def\iccvPaperID{5158} 

\ificcvfinal\fi

\begin{document}

\title{Seeing What a GAN Cannot Generate}

\author{%
David Bau\textsuperscript{1,2},
Jun-Yan Zhu\textsuperscript{1},
Jonas Wulff\textsuperscript{1},
William Peebles\textsuperscript{1} \\
Hendrik Strobelt\textsuperscript{2},
Bolei Zhou\textsuperscript{3},
Antonio Torralba\textsuperscript{1,2} \\
\textsuperscript{1}MIT CSAIL,
\textsuperscript{2}MIT-IBM Watson AI Lab,
\textsuperscript{3}The Chinese University of Hong Kong
}

\maketitle
\thispagestyle{empty}

\begin{abstract}
Despite the success of Generative Adversarial Networks (GANs), mode collapse remains a serious issue during GAN training. To date, little work has focused on understanding and quantifying which modes have been dropped by a model. In this work, we visualize mode collapse at both the distribution level and the instance level. First, we deploy a semantic segmentation network to compare the distribution of segmented objects in the generated images with the target distribution in the training set. Differences in statistics reveal object classes that are omitted by a GAN.  Second, given the identified omitted object classes, we visualize the GAN's omissions directly. In particular, we compare specific differences between individual photos and their approximate inversions by a GAN. To this end, we relax the problem of inversion and solve the tractable problem of inverting a GAN layer instead of the entire generator. Finally, we use this framework to analyze several recent GANs trained on multiple datasets and identify their typical failure cases.
\end{abstract}

\section{Introduction}
\lblsec{intro}
\noindent
The remarkable ability of a Generative Adversarial Network (GAN) to synthesize realistic images leads us to ask: How can we know what a GAN is \textit{unable} to generate? Mode-dropping or mode collapse, where a GAN omits portions of the target distribution, is seen as one of the biggest challenges for GANs~\citep{goodfellow2016nips,li2018implicit}, yet current analysis tools provide little insight into this phenomenon in state-of-the-art GANs.

Our paper aims to provide detailed insights about dropped modes.  Our goal is not to measure GAN quality using a single number: existing metrics such as Inception scores~\citep{salimans2016improved} and \fid~\citep{heusel2017gans} focus on that problem.  While those numbers measure \emph{how far} the generated and target distributions are from each other, we instead seek to understand \emph{what} is different between real and fake images.  Existing literature typically answers the latter question by sampling generated outputs, but such samples only visualize what a GAN is \emph{capable} of doing.  We address the complementary problem: we want to see what a GAN \emph{cannot} generate.
\begin{figure}
\centering
\ifdefined\iccv
\includegraphics[width=\columnwidth]{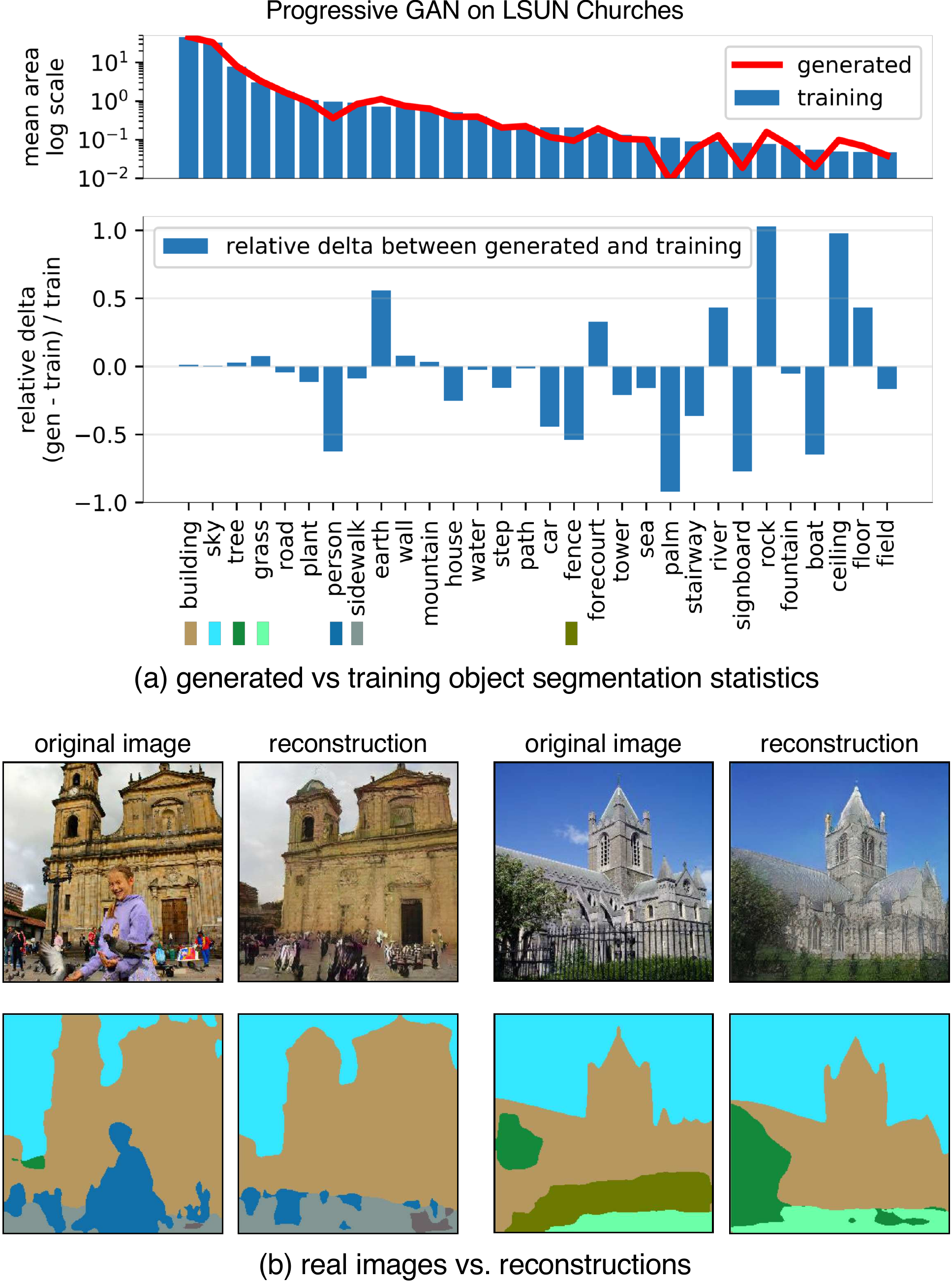}%
\else
\includegraphics[width=\columnwidth]{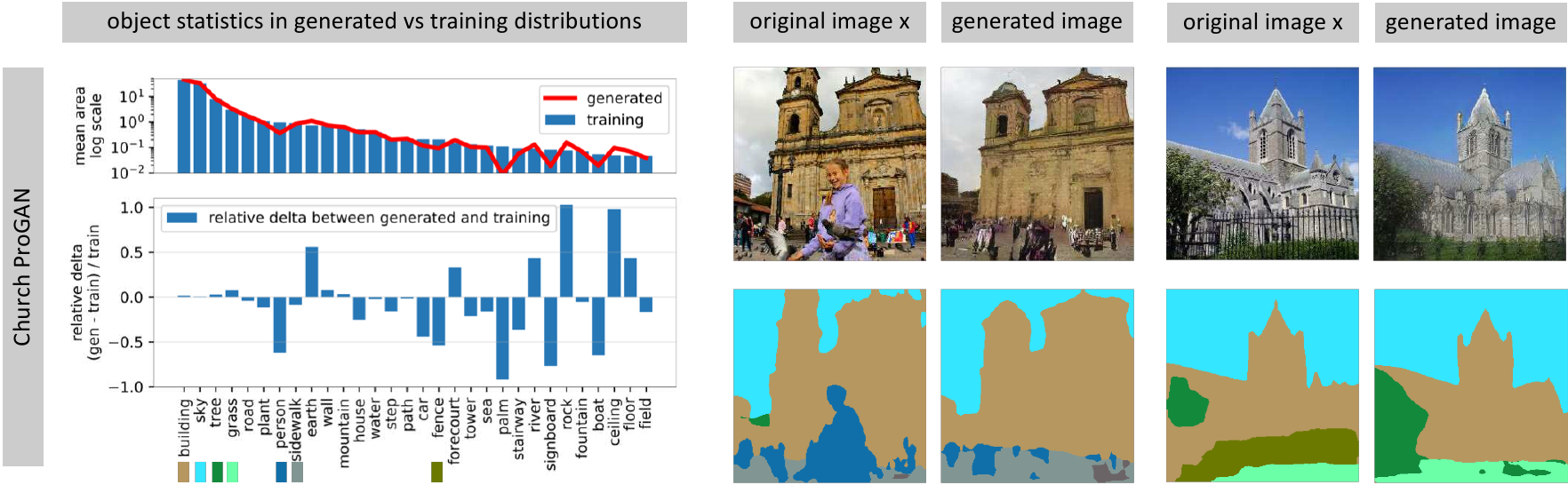}%
\vspace{-0.1in}%
\fi
\caption{Seeing what a GAN cannot generate: (a) We compare the distribution of object segmentations in the training set of LSUN churches~\cite{yu2015lsun} to the distribution in the generated results: objects such as people, cars, and fences are dropped by the generator. (b) We compare pairs of a real image and its reconstruction  in which individual instances of a person and a fence cannot be generated.  In each block, we show a real photograph (top-left), a generated reconstruction (top-right), and segmentation maps for both (bottom).}
\ifdefined\iccv
\vspace{-0.2in}
\else
\vspace{-0.3in}
\fi
\lblfig{teaser}
\end{figure}

In particular, we wish to know: Does a GAN deviate from the target distribution by ignoring difficult images altogether? Or are there specific, semantically meaningful parts and objects that a GAN decides not to learn about? And if so, how can we detect and visualize these missing concepts that a GAN does not generate?

Image generation methods are typically tested on images of faces, objects, or scenes. Among these, scenes are an especially fertile test domain as each image can be parsed into clear semantic components by segmenting the scene into objects. Therefore, we propose to directly understand mode dropping by analyzing a scene generator at two levels: the distribution level and instance level.

First, we characterize omissions in the distribution as a whole, using \textit{\gissfull}: we segment both generated and ground truth images and compare the distributions of segmented object classes.
For example, \reffig{teaser}a shows that in a church GAN model, object classes such as people, cars, and fences appear on fewer pixels of the generated distribution as compared to the training distribution.

Second, once omitted object classes are identified, we want to visualize specific examples of failure cases.
To do  so, we must find image instances where the GAN \textit{should} generate an object class but does not.
We find such cases using a new reconstruction method called \textit{Layer Inversion} which relaxes reconstruction to a tractable problem. Instead of inverting the entire GAN, our method inverts a layer of the generator.  Unlike existing methods to invert a small generator~\citep{zhu2016generative,brock2017neural}, our method allows us to create reconstructions for complex, state-of-the-art GANs. 
Deviations between the original image and its reconstruction reveal image features and objects that the generator cannot draw faithfully.

We apply our framework to analyze several recent GANs trained on different scene datasets.
Surprisingly, we find that dropped object classes are not distorted or rendered in a low quality or as noise. Instead, they are simply not rendered at all, \textit{as if the object was not part of the scene}. 
For example, in \reffig{teaser}b, we observe that large human figures are skipped entirely, and the parallel lines in a fence are also omitted.
Thus a GAN can ignore classes that are too hard, while at the same time producing outputs of high average visual quality. 
Code, data, and additional information are available at \textbf{\href{http://ganseeing.csail.mit.edu}{ganseeing.csail.mit.edu}}.


\section{Related work}
\vspace{5pt}\myparagraph{Generative Adversarial Networks~\citep{goodfellow2014generative}} have enabled many computer vision and graphics applications such
as generation~\cite{brock2019large,karras2018progressive,karras2019style}, image and video manipulation~\cite{huang2018multimodal,isola2017image,park2019SPADE,sangkloy2016scribbler,taigman2017unsupervised,wang2018vid2vid,zhu2017unpaired}, object recognition~\cite{bousmalis2017unsupervised,wang2017fast}, and text-to-image translation~\cite{reed2016generative,xu2017attngan,zhang2017stackgan}. One important issue in this emerging topic is how to evaluate and compare different methods~\cite{theis2016note,wu2016quantitative}. For example, many evaluation metrics have been proposed to evaluate unconditional GANs such as Inception score~\cite{salimans2016improved}, \fid~\cite{heusel2017gans}, and \wsd~\cite{karras2018progressive}. Though the above metrics can quantify different aspects of model performance, they cannot explain what visual content the models fail to synthesize. Our goal here is \emph{not} to introduce a metric. Instead, we aim to provide explanations of a common failure case of GANs: mode collapse. Our error diagnosis tools complement existing single-number metrics and can provide additional insights into the model's limitations.

\myparagraph{Network inversion.}
Prior work has found that inversions of GAN generators are useful for photo manipulation~\citep{bau2019semantic,brock2017neural,peleg2018structured,zhu2016generative} and unsupervised feature learning~\cite{donahue2016adversarial,dumoulin2016adversarially}. Later work found that DCGAN left-inverses can be computed to high precision~\cite{lipton2017precise,yeh2017semantic}, and that inversions of a GAN for glyphs can reveal specific strokes that the generator is unable to generate~\cite{creswell2018inverting}.  While previous work~\cite{zhu2016generative} has investigated inversion of 5-layer DCGAN generators, we find that when moving to a 15-layer Progressive GAN, high-quality inversions are much more difficult to obtain. In our work, we develop a layer-wise inversion method that is more effective for these large-scale GANs. We apply a classic layer-wise training approach~\cite{bengio2007greedy,hinton2006reducing} to the problem of training an encoder and further introduce layer-wise image-specific optimization.
Our work is also loosely related to inversion methods for understanding CNN features and classifiers~\cite{dosovitskiy2016inverting,mahendran2015understanding,olah2017feature,olah2018building}. However, we focus on understanding generative models rather than classifiers. 

\myparagraph{Understanding and visualizing networks.} Most prior work on network visualization concerns discriminative classifiers~\cite{bach2015pixel,bau2017network,kindermans2017reliability,lundberg2017unified,smilkov2017smoothgrad,springenberg2014striving,zeiler2014visualizing,zhou2014learning}. GANs have been visualized by examining the discriminator~\cite{radford2015unsupervised} and the semantics of internal features~\cite{bau2019gandissect}. Different from recent work~\cite{bau2019gandissect} that aims to understand what a GAN has learned, our work provides a complementary perspective and focuses on 
what semantic concepts a GAN fails to capture.

 


\section{Method}
\lblsec{methods}
\noindent
Our goal is to visualize and understand the semantic concepts that a GAN generator cannot generate, in both the entire distribution and in each image instance. We will proceed in two steps.  First, we measure \textit{\gissfull} by segmenting both generated and target images and identifying types of objects that a generator omits when compared to the distribution of real images. 
Second, we visualize how the dropped object classes are omitted for individual images by finding real images that contain the omitted classes and projecting them to their best reconstruction given an intermediate layer of the generator.
We call the second step \textit{Layer Inversion}.

\subsection{Quantifying distribution-level mode collapse}
\noindent
The systematic errors of a GAN can be analyzed by exploiting the hierarchical structure of a scene image. Each scene has a natural decomposition into objects, so we can estimate deviations from the true distribution of scenes by estimating deviations of constituent object statistics.  For example, a GAN that renders bedrooms should also render some amount of curtains. If the curtain statistics depart from what we see in true images, we will know we can look at curtains to see a specific flaw in the GAN.

To implement this idea, we segment all the images using the Unified Perceptual Parsing network~\citep{xiao2018unified}, which labels each pixel of an image with one of $336$ object classes. Over a sample of images, we measure the total area in pixels for each object class and collect mean and covariance statistics for all segmented object classes. We sample these statistics over a large set of generated images as well as training set images. We call the statistics over all object segmentations \textit{\gissfull}.

\begin{figure*}
\centering
\includegraphics[width=\textwidth]{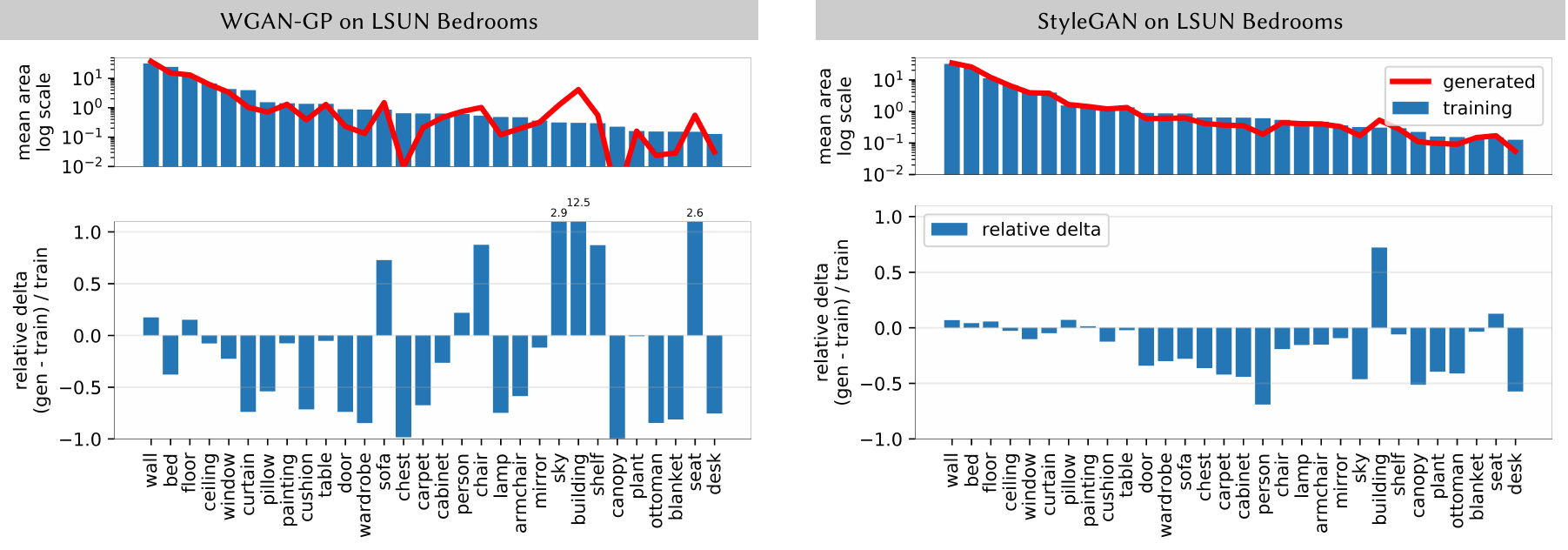}
\vspace{-0.1in}
\caption{Using \gissfull to understand the different behavior of the two models trained on LSUN bedrooms~\cite{yu2015lsun}.  The histograms reveal that WGAN-GP~\cite{gulrajani2017improved} (left) deviates from the true distribution much more than StyleGAN~\cite{karras2019style} (right), identifying segmentation classes that are generated too little and others that are generated too much. For example, WGAN-GP does not generate enough pixels containing beds, curtains, or cushions compared to the true distribution of bedroom images, while StyleGAN correctly matches these statistics. StyleGAN is still not perfect, however, and does not generate enough doors, wardrobes, or people. Numbers above bars indicate clipped values beyond the range of the chart.}
\lblfig{comparing-stats}
\vspace{-0.15in}
\end{figure*}
\reffig{comparing-stats} visualizes mean statistics for two networks.  In each graph, the mean segmentation frequency for each generated object class is compared to that seen in the true distribution. Since most classes do not appear on most images, we focus on the most common classes by sorting classes by descending frequency. The comparisons can reveal many specific differences between recent state-of-the-art models.  Both analyzed models are trained on the same image distribution (LSUN bedrooms~\cite{yu2015lsun}), but WGAN-GP~\cite{gulrajani2017improved} departs from the true distribution much more than StyleGAN~\cite{karras2019style}.  

It is also possible to summarize statistical differences in segmentation in a single number.  To do this, we define the \fssfull (\fssshort), which is an interpretable analog to the popular \fid(FID) metric~\citep{heusel2017gans}:
$\text{\fssshort} \equiv || \mu_g - \mu_t ||^2 + \Tr( \Sigma_g + \Sigma_t - 2(\Sigma_g \Sigma_t)^{1/2})$. 
In our \fssshort formula, $\mu_t$ is the mean pixel count for each object class over a sample of training images, and $\Sigma_t$ is the covariance of these pixel counts.  Similarly, $\mu_g$ and $\Sigma_g$ reflect segmentation statistics for the generative model. In our experiments, we compare statistics between $10{,}000$ generated samples and $10{,}000$ natural images.

\gissfull measure the entire distribution: for example, they reveal when a generator omits a particular object class. However, they do not single out specific images where an object should have been generated but was not. To gain further insight, we need a method to visualize omissions of the generator for each image. 

\subsection{Quantifying instance-level mode collapse}
\lblsec{method-layer-wise}
\noindent
To address the above issue, we compare image pairs $(\rvx, \rvx')$, where $\rvx$ is a real image that contains a particular object class dropped by a GAN generator $G$, and $\rvx'$ is a projection onto the space of all images that can be generated by a layer of the GAN model. 

\myparagraph{Defining a tractable inversion problem.} In the ideal case, we would like to find an image that can be perfectly synthesized by the generator $G$ and stay close to the real image $\rvx$. Formally, we seek  $\rvx' = G(\rvz^*)$, where  $\rvz^* = \argmin_{\rvz} \disti(G(\rvz), \rvx)$ and $\disti$ is a distance metric in image feature space. 
Unfortunately, as shown in \refsec{results-prior-inversion}, previous methods~\cite{donahue2016adversarial,zhu2016generative} fail to solve this full inversion problem for recent generators due to the large number of layers in $G$.  Therefore, we instead solve a tractable subproblem of full inversion.  We decompose the generator $G$ into layers
\begin{align}
G = G_f(g_n(\cdots((g_1(\rvz)))),
\end{align}
where $g_1,..., g_n$ are several early layers of the generator, and $G_f$ groups all the \final layers of the $G$ together.

Any image that can be generated by $G$ can also be generated by $G_f$. That is, if we denote by $\range(G)$ the set of all images that can be output by $G$, then we have
$\range(G) \subset \range(G_f)$.  That implies, conversely, that any image that cannot be generated by $G_f$ cannot be generated by $G$ either.  Therefore any omissions we can identify in $\range(G_f)$ will also be omissions of $\range(G)$.

Thus for layer inversion, we visualize omissions by solving the easier problem of inverting the \final layers $G_f$:
\begin{align}
\rvx' & = G_f(\rvr^*), \\ \nonumber
\text{where } \rvr^* & = \argmin_{\rvr} \disti(G_f(\rvr), \rvx).
\end{align}

\begin{figure}[!t]
\centering
\includegraphics[width=\linewidth]{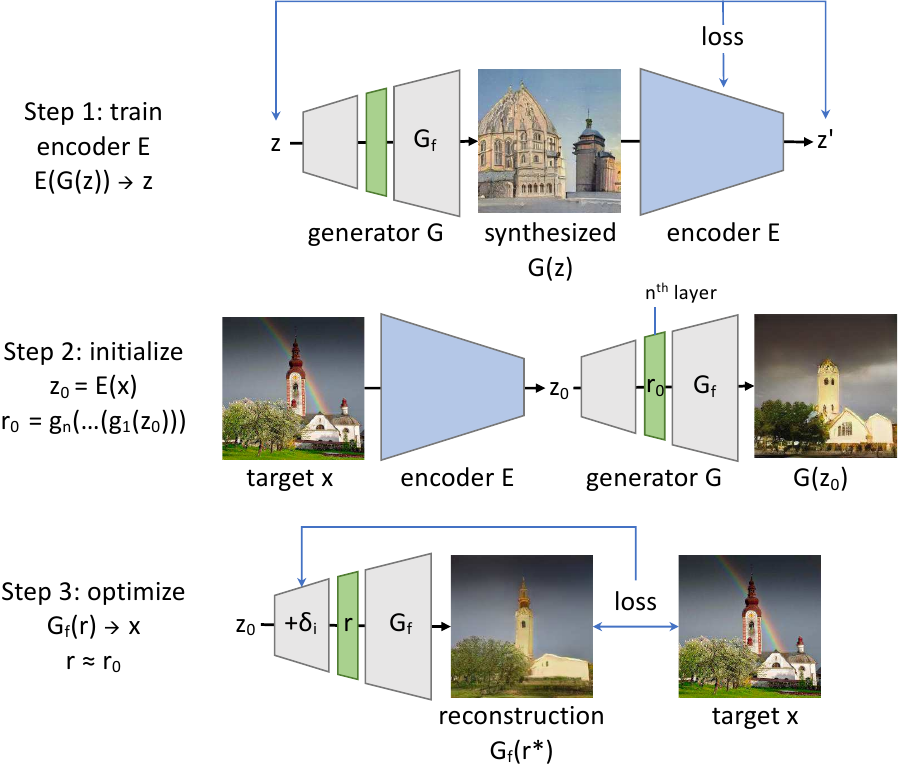}
\caption{Overview of our layer inversion method.  First, we train a network $E$ to invert $G$; this is used to obtain an initial guess of the latent  $\rvz_0=E(\rvx)$ and its intermediate representation $\rvr_0 = g_n(\cdots(g_1(\rvz_0)))$.  Then $\rvr_0$ is used to initialize a search for $\rvr^*$ to obtain a reconstruction $\rvx' = G_f(\rvr^*)$ close to the target $\rvx$.}
\label{fig:framework_inversion}
\end{figure}

Although we ultimately seek an intermediate representation $\rvr$, it will be helpful to begin with an estimated $\rvz$: an initial guess for $\rvz$ helps us regularize our search to favor values of $\rvr$ that are more likely to be generated by a $\rvz$.  Therefore, we solve the inversion problem in two steps: First we construct a neural network $E$ that approximately inverts the entire $G$ and computes an estimate $\rvz_0 = E(\rvx)$. Subsequently we solve an optimization problem to identify an intermediate representation $\rvr^* \approx \rvr_0 = g_n(\cdots(g_1(\rvz_0)))$ that generates a reconstructed image $G_f(\rvr^*)$ to closely recover $\rvx$. \reffig{framework_inversion} illustrates our layer inversion method.

\myparagraph{Layer-wise network inversion.}
A deep network can be trained more easily by pre-training individual layers on smaller problems~\cite{hinton2006reducing}.  Therefore, to learn the inverting neural network $E$, we also proceed layer-wise.  For each layer $g_i \in \{g_1,...,g_n, G_f\}$, we train a small network $e_i$ to approximately invert $g_i$. That is, defining $\rvr_i = g_i(\rvr_{i-1})$, our goal is to learn a network $e_i$ that approximates the computation $\rvr_{i-1} \approx e_i(\rvr_{i})$. 
We also want the predictions of the network $e_i$ to well preserve the output of the layer $g_i$, so we want $\rvr_{i} \approx g_i(e_i(\rvr_{i}))$.
We train $e_i$ to minimize both left- and right-inversion losses:
\begin{align}
\gL_{\text{L}} & \equiv \E_{\rvz}[||\rvr_{i-1}- e(g_i(\rvr_{i-1}))||_1] \nonumber\\
\gL_{\text{R}} & \equiv \E_{\rvz}[||\rvr_i - g_i(e(\rvr_i))||_1] \nonumber \\
e_i & = \argmin_{e} \quad \gL_{\text{L}} + \lambda_{\text{R}} \gL_{\text{R}},
\end{align}
To focus on training near the manifold of representations produced by the generator, we sample $\rvz$ and then use the layers $g_i$ to compute samples of $\rvr_{i-1}$ and $\rvr_i$, so $\rvr_{i-1} = g_{i-1}(\cdots g_1(\rvz) \cdots)$.
Here $||\cdot||_1$ denotes an L1 loss, and we set $\lambda_R = 0.01$ to emphasize the reconstruction of $\rvr_{i-1}$.

Once all the layers are inverted, we can compose an inversion network for all of $G$:
\begin{align}
E^* & = e_1(e_2(\cdots(e_n(e_f(\rvx))))).
\end{align}
The results can be further improved by jointly fine-tuning this composed network $E^*$ to invert $G$ as a whole. We denote this fine-tuned result as $E$.

\myparagraph{Layer-wise image optimization.}


As described at the beginning of \refsec{method-layer-wise}, inverting the entire $G$ is difficult: $G$ is non-convex, and optimizations over $\rvz$ are quickly trapped in local minima. Therefore, after obtaining a decent initial guess for $\rvz$, we turn our attention to the more relaxed optimization problem of inverting the layers $G_f$; that is, starting from $\rvr_0 = g_n(\cdots(g_1(\rvz_0)))$, we seek an intermediate representation $\rvr^*$ that generates a reconstructed image $G_f(\rvr^*)$ to closely recover $\rvx$.

To regularize our search to favor $\rvr$ that are close to the representations computed by the early layers of the generator, we search for $\rvr$ that can be computed by making small perturbations of the early layers of the generator:
\begin{align}
    \rvz_0 & \equiv E(\rvx) \nonumber \\
    \rvr & \equiv \delta_n + g_n(\cdots (\delta_2 + g_2(\delta_1 + g_1(\rvz_0)))) \nonumber \\
    \rvr^* & = \argmin_{\rvr} \left( \dist(\rvx, G_{f}(\rvr)) + \lambda_{\text{reg}} \sum_i || \delta_i ||^2 \right).
\end{align}
That is, we begin with the guess $\rvz_0$ given by the neural network $E$, and then we learn small perturbations of each layer before the $n$-th layer, to obtain an $\rvr$ that reconstructs the image $\rvx$ well. For $\dist$ we sum image pixel loss and VGG perceptual loss~\cite{simonyan2014very}, similar to existing reconstruction methods~\cite{dosovitskiy2016inverting,zhu2016generative}. 
The hyper-parameter $\lambda_{\text{reg}}$ determines the balance between image reconstruction loss and the regularization of $\rvr$. We set $\lambda_{\text{reg}} = 1$ in our experiments.

\section{Results}
\lblsec{results}

\begin{figure}
\centering
\vspace{-2pt}
\includegraphics[width=0.92\columnwidth]{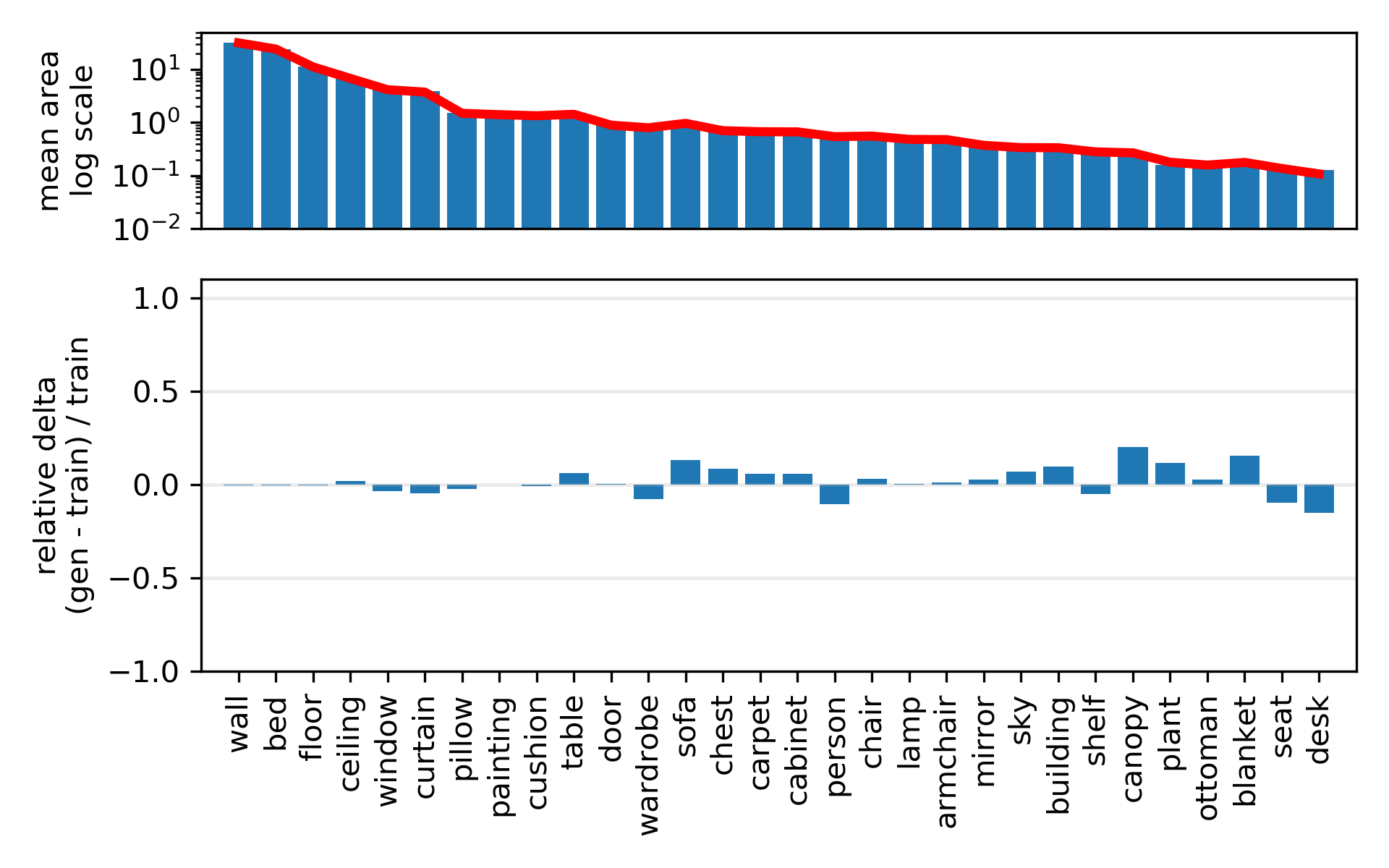}%
\vspace{-5pt}%
\caption{Sensitivity test for \gissfull.  This plot compares two different random samples of $10,000$ images from the LSUN bedroom dataset.  An infinite-sized sample would show no differences; the observed differences reveal the small measurement noise introduced by the finite sampling process.}
\lblfig{zero-baseline}
\end{figure}
\begin{figure}
\centering
\ifdefined\iccv
\includegraphics[width=\columnwidth]{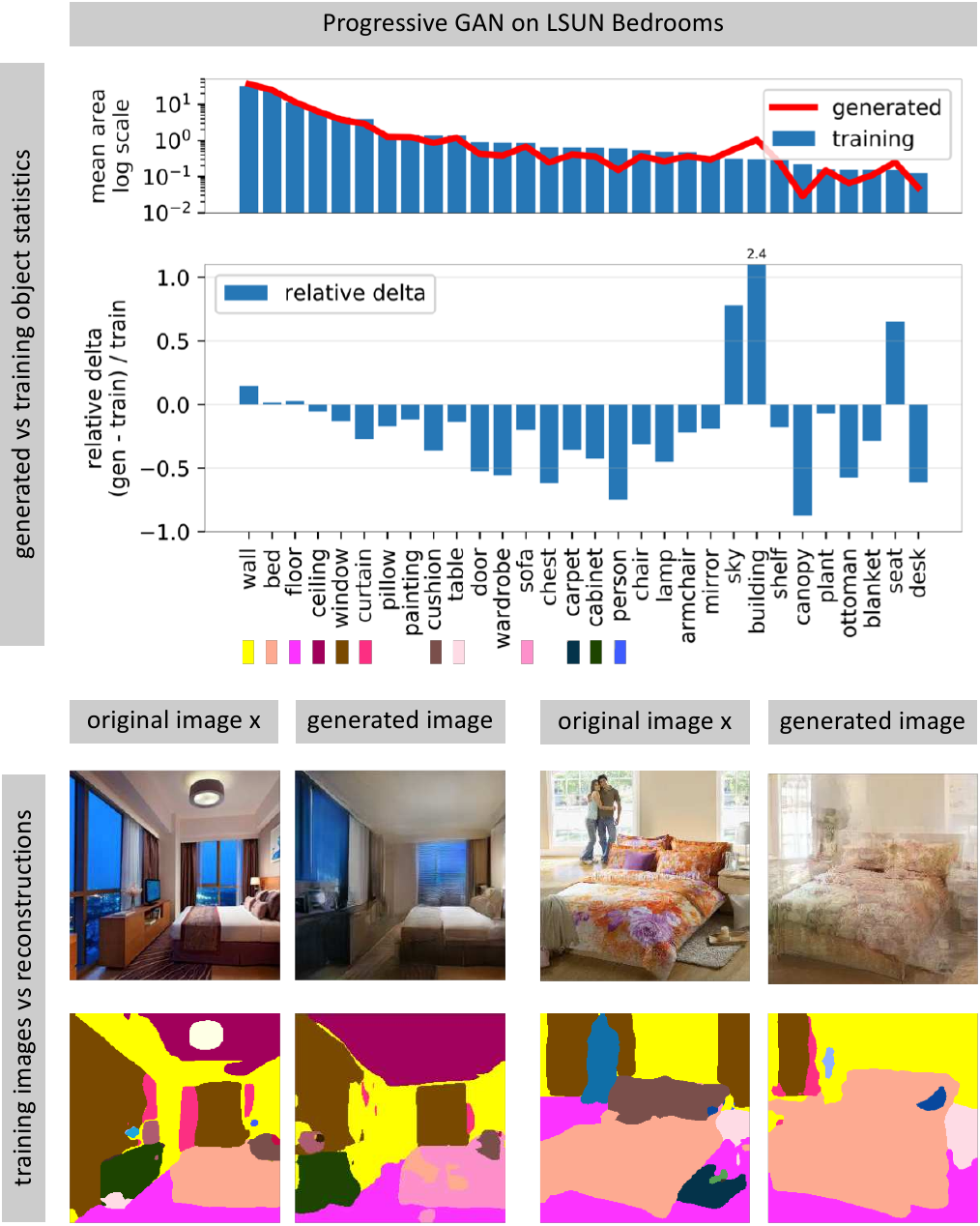}
\else
\includegraphics[width=\columnwidth]{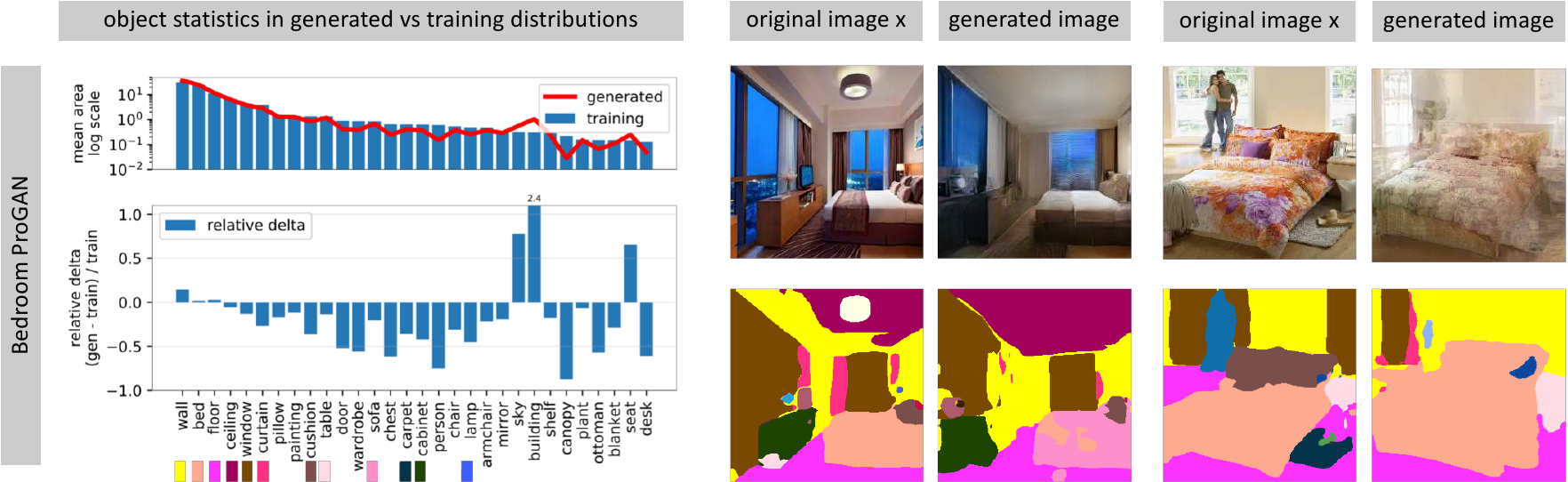}
\fi
\caption{A visualization of the omissions of a bedroom generator; a Progressive GAN for LSUN bedrooms is tested.  On top, a comparison of object distributions shows that many classes of objects are left out by the generator, including people, cushions, carpets, lamps, and several types of furniture.  On the bottom, photographs are shown with their reconstructions $G(E(\rvx))$, along with segmentations.  These examples directly reveal many object classes omitted by the bedroom generator.}
\lblfig{bedroom}
\end{figure}

\noindent \textbf{Implementation details.} We analyze three recent models: WGAN-GP~\citep{gulrajani2017improved}, Progressive GAN~\citep{karras2018progressive}, and StyleGAN~\citep{karras2019style}, trained on LSUN bedroom images~\citep{yu2015lsun}. In addition, for Progressive GAN we analyze a model trained to generate LSUN church images.  To segment images, we use the Unified Perceptual Parsing network~\citep{xiao2018unified}, which labels each pixel of an image with one of $336$ object classes.  Segmentation statistics are computed over samples of $10{,}000$ images.

\subsection{\gissfull}
\noindent
We first examine whether segmentation statistics correctly reflect the output quality of models across architectures. \reffig{comparing-stats} and \reffig{bedroom}  show \gissfull for WGAN-GP~\citep{gulrajani2017improved},  StyleGAN~\citep{karras2019style}, and Progressive GAN~\citep{karras2018progressive} trained on LSUN bedrooms~\citep{yu2015lsun}. The histograms reveal that, for a variety of segmented object classes, StyleGAN matches the true distribution of those objects better than Progressive GAN, while WGAN-GP matches least closely.  The differences can be summarized using \fssfull (\reftbl{FSD}), confirming that better models match the segmented statistics better overall.

\begin{table}[h!]%
\centering\vspace{-5pt}%
\begin{tabular}{lccr} 
   \textbf{Model} &  \textbf{\fssshort}  \\
   \midrule %
   WGAN-GP~\citet{gulrajani2017improved} bedrooms (\reffig{comparing-stats}) & 428.4 \\
   ProGAN~\citet{karras2018progressive} bedrooms (\reffig{bedroom}) & 85.2 \\
   StyleGAN~\citet{karras2019style} bedrooms (\reffig{comparing-stats}) & 22.6
\end{tabular}\vspace{5pt}
\caption{\fssshort summarizes \gissfull}
\lbltbl{FSD}
\end{table}

\subsection{Sensitivity test}
\noindent
\reffig{zero-baseline} illustrates the sensitivity of measuring \gissfull over a finite sample of $10{,}000$ images.  Instead of comparing a GAN to the true distribution, we compare two different randomly chosen subsamples of the LSUN bedroom data set to each other.  A perfect test with infinite sample sizes would show no difference; the small differences shown reflect the sensitivity of the test and are due to sampling error.

\subsection{Identifying dropped modes}
\noindent
\reffig{teaser} and \reffig{bedroom} show the results of applying our method to analyze the generated segmentation statistics for Progressive GAN models of churches and bedrooms.  Both the histograms and the instance visualizations provide insight into the limitations of the generators.

The histograms reveal that the generators partially skip difficult subtasks.  For example, neither model renders as many people as appear in the target distribution.
We use inversion to create reconstructions of natural images that include many pixels of people or other under-represented objects. \reffig{teaser} and \reffig{bedroom} each shows two examples on the bottom. Our inversion method reveals the way in which the models fail.  The gaps are not due to low-quality rendering of those object classes, but due to the wholesale omission of these classes.  For example, large human figures and certain classes of objects are not included.


\begin{figure*}
\centering
\includegraphics[width=\textwidth]{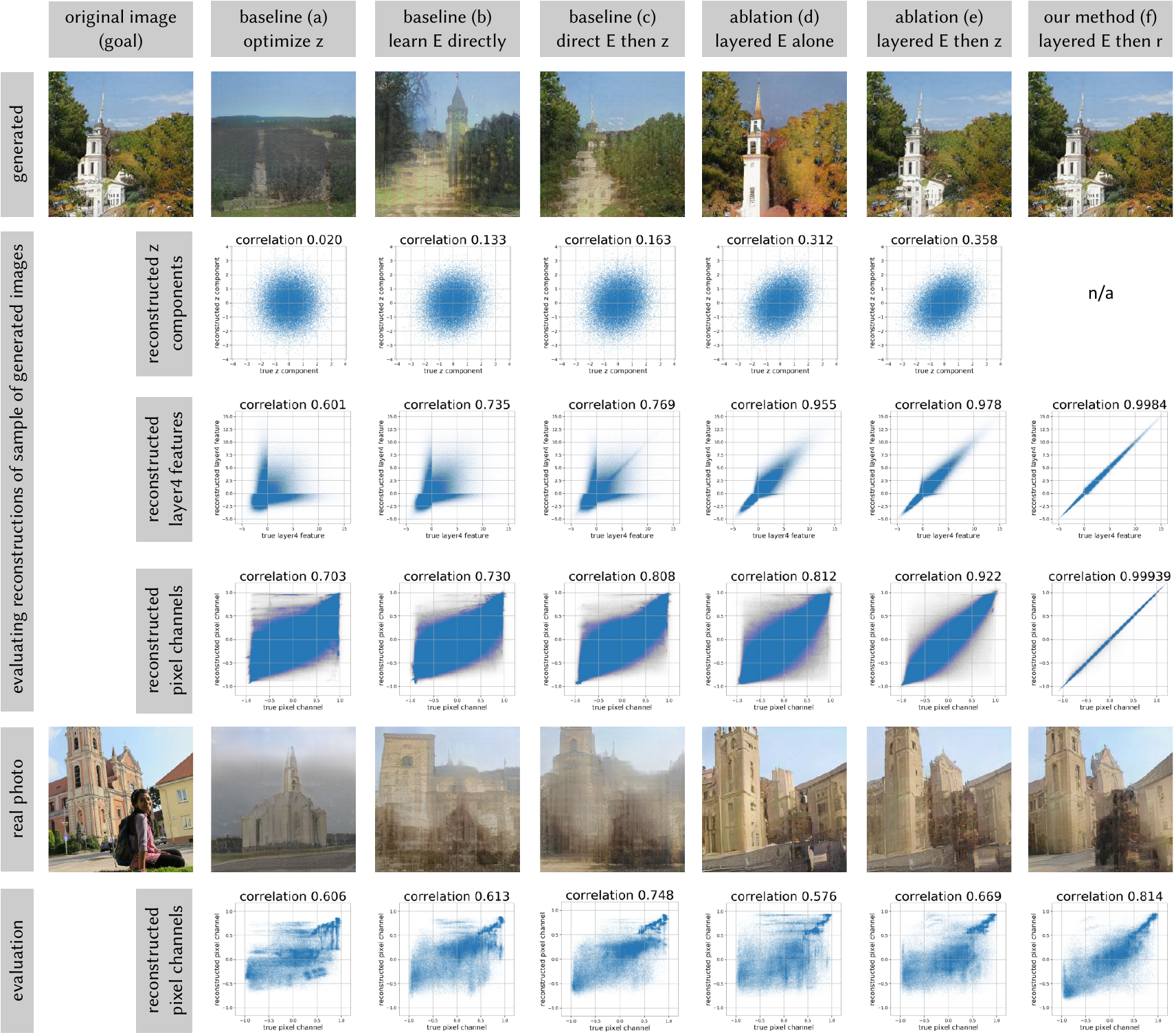}
\vspace{-0.18in}
\caption{Comparison of methods to invert the generator of Progressive GAN trained to generate LSUN church images. Each method is described; (a) (b) and (c) are baselines, and (d), (e), and (f) are variants of our method. The first four rows show behavior given GAN-generated images as input. In the scatter plots, every point plots a reconstructed component versus its true value, with a point for every RGB pixel channel or every dimension of a representation.  Reconstruction accuracy is shown as mean correlation over all dimensions for $\rvz$, \texttt{layer4}, and image pixels, based on a sample of $100$ images.  Our method (f) achieves nearly perfect reconstructions of GAN-generated images. In the bottom rows, we apply each of the methods on a natural image.
}
\lblfig{method-grid}
\end{figure*}

\subsection{Layer-wise inversion vs other methods}
\lblsec{results-prior-inversion}
\noindent
We compare our layer-wise inversion method to several previous approaches; we also benchmark it against ablations of key components of the method.

The first three columns of \reffig{method-grid} compare our method to prior inversion methods. We test each method on a sample of $100$ images produced by the generator $G$, where the ground truth $\rvz$ is known, and the reconstruction of an example image is shown.  In this case an ideal inversion should be able to perfectly reconstruct $\rvx' = \rvx$. In addition, a reconstruction of a real input image is shown at the bottom.  While there is no ground truth latent and representation for this image, the qualitative comparisons are informative.

\myparagraph{(a) Direct optimization of $\rvz$.}  Smaller generators such as 5-layer DCGAN~\cite{radford2015unsupervised} can be inverted by applying gradient descent on $\rvz$ to minimize reconstruction loss~\citep{zhu2016generative}.
In column (a), we test this method on a 15-layer Progressive GAN and find that neither $\rvz$ nor $\rvx$ can be constructed accurately.

\myparagraph{(b): Direct learning of $E$.}  Another natural solution~\cite{donahue2016adversarial,zhu2016generative} is to learn a deep network $E$ that inverts $G$ directly, without the complexity of layer-wise decomposition.
Here, we learn an inversion network with the same parameters and architecture as the network $E$ used in our method, but train it end-to-end by directly minimizing expected reconstruction losses over generated images, rather than learning it by layers.  The method does benefit from the power of a deep network to learn generalized rules~\cite{gershman2014amortized}, and the results are marginally better than the direct optimization of $\rvz$.  However, both qualitative and quantitative results remain poor.

\myparagraph{(c): Optimization of $\rvz$ after initializing with $E(\rvx)$.}  This is the full method used in \citet{zhu2016generative}.  By initializing method (a) using an initial guess from method (b), results can be improved slightly.  For smaller generators, this method performs better than method (a) and (b).  However, when applied to a Progressive GAN, the reconstructions are far from satisfactory.

\myparagraph{Ablation experiments.} The last three columns of \reffig{method-grid} compare our full method (f) to two ablations of our method.

\myparagraph{(d): Layer-wise network inversion only.}
We can simply use the layer-wise-trained inversion network $E$ as the full inverse, and simply use the initial guess $\rvz_0 = E(\rvx)$, setting $\rvx' = G(\rvz_0)$.  This fast method requires only a single forward pass through the inverter network $E$.  The results are better than the baseline methods but far short of our full method. 

Nevertheless, despite the inaccuracy of the latent code $\rvz_0$, the intermediate layer features are highly correlated with their true values; this method achieves $95.5\%$ correlation versus the true $\rvr_4$.  Furthermore, the qualitative results show that when reconstructing real images, this method obtains more realistic results despite being noticeably different from the target image.

\myparagraph{(e): Inverting $G$ without relaxation to $G_f$.}  We can improve the initial guess $\rvz_0=E(\rvx)$ by directly optimizing $\rvz$ to minimize the same image reconstruction loss. This marginally improves upon $\rvz_0$. However, the reconstructed images and the input images still differ signficantly, and the recovery of $\rvz$ remains poor.  Although the qualitative results are good, the remaining error means that we cannot know if any reconstruction errors are due to failures of $G$ to generate an image, or if those reconstruction errors are merely due to the inaccuracy of the inversion method.

\begin{figure*}[t]
\vspace{-5pt}%
\centering
\includegraphics[width=\textwidth]{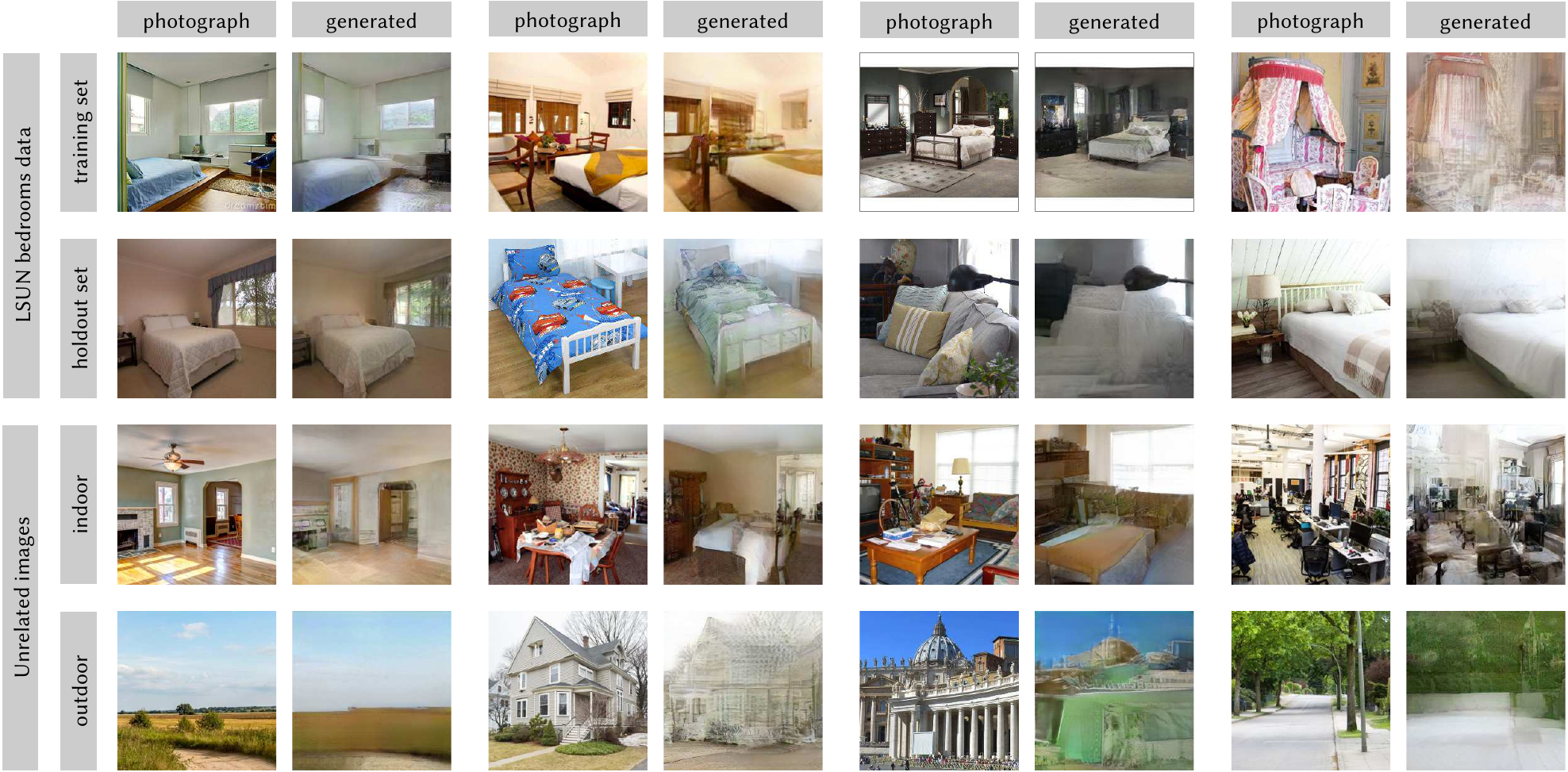}
\caption{Inverting layers of a Progressive GAN bedroom generator. From top to bottom:  uncurated reconstructions of photographs from the LSUN training set, the holdout set, and unrelated (non-bedroom) photographs, both indoor and outdoor.}
\lblfig{bedroom-ex}
\vspace{-5pt}
\end{figure*}

\myparagraph{(f): Our full method.}  By relaxing the problem and regularizing optimization of $\rvr$ rather than $\rvz$, our method achieves nearly perfect reconstructions of both intermediate representations and pixels.  Denote the full method as $\rvr^* = E_f(\rvx)$.


The high precision of $E_f$ within the range of $G$ means that, when we observe large differences between  $\rvx$ and $G_f(E_f(\rvx))$,  they are unlikely to be a failure of $E_f$.  This indicates that $G_f$ cannot render $\rvx$, which means that $G$ cannot either. Thus our ability to solve the relaxed inversion problem with an accuracy above $99\%$ gives us a reliable tool to visualize samples that reveal what $G$ cannot do.

Note that the purpose of $E_f$ is to show dropped modes, not positive capabilities. The range of $G_f$ upper-bounds the range of $G$, so the reconstruction $G_f(E_f(\rvx))$ could be better than what the full network $G$ is capable of.  For a more complete picture, methods (d) and (e) can be additionally used as lower-bounds: those methods do not prove images are outside $G$'s range, but they can reveal positive capabilities of $G$ because they construct generated samples in $\range(G)$.
\begin{figure*}[t]
\centering
\includegraphics[width=\textwidth]{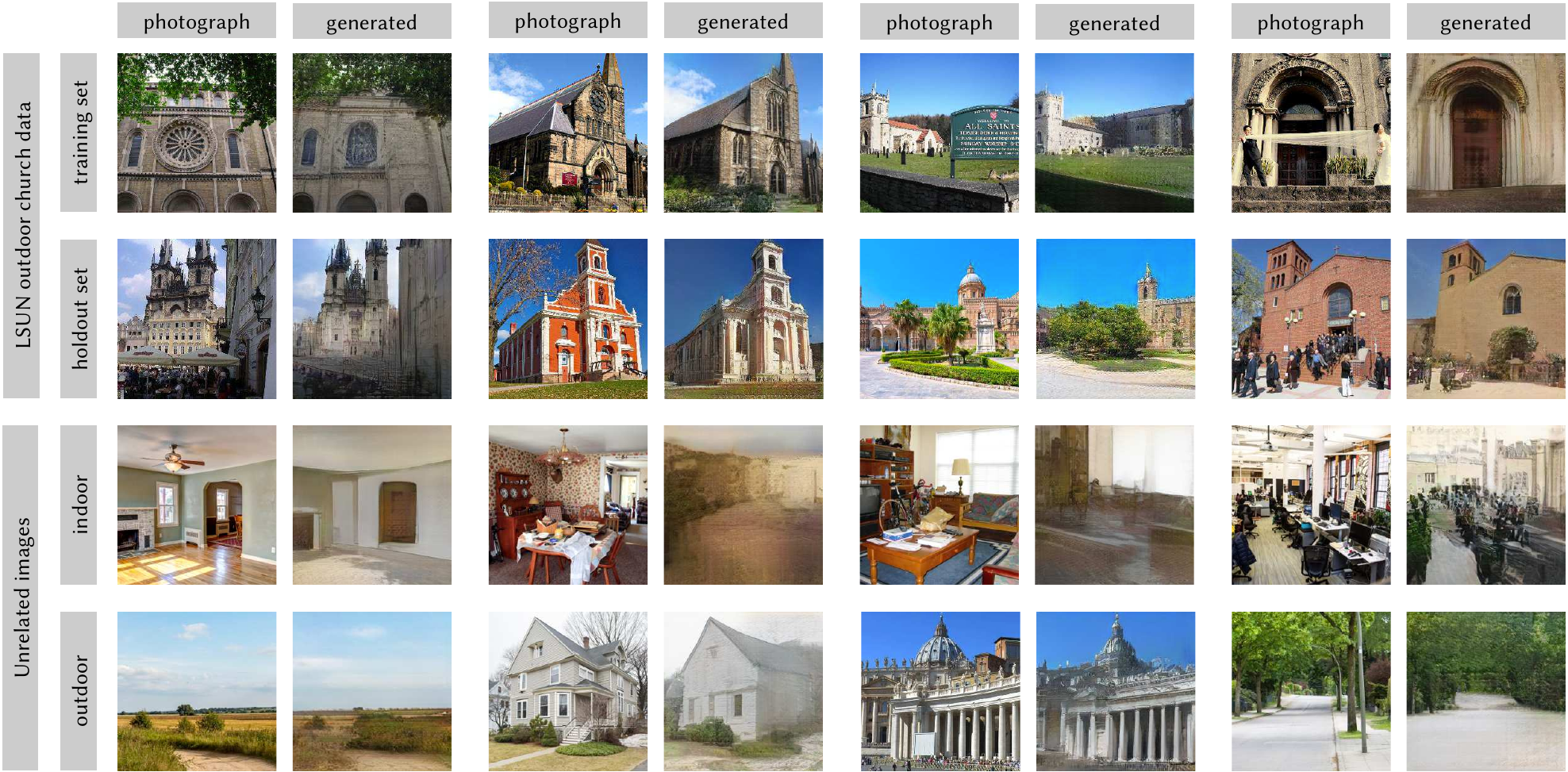}
\caption{Inverting layers of a Progressive GAN outdoor church generator. From top to bottom: uncurated reconstructions of photographs from the LSUN training set, the holdout set, and unrelated (non-church) photographs, both indoor and outdoor.}
\vspace{-5pt}
\lblfig{church-ex}
\end{figure*}

\subsection{Layer-wise inversion across domains}
\noindent
Next, we apply the inversion tool to test the ability of generators to synthesize images outside their training sets.  \reffig{bedroom-ex} shows qualitative results of applying method (f) to invert and reconstruct natural photographs of different scenes using a Progressive GAN trained to generate LSUN bedrooms.  Reconstructions from the LSUN training and LSUN holdout sets are shown; these are compared to newly collected unrelated (non-bedroom) images taken both indoors and outdoors.  Objects that disappear from the reconstructions reveal visual concepts that cannot be represented by the model.  Some indoor non-bedroom images are rendered in a bedroom style: for example, a dining room table with a white tablecloth is rendered to resemble a bed with a white bed sheet. As expected, outdoor images are not reconstructed well.

\reffig{church-ex} shows similar qualitative results using a Progressive GAN for LSUN outdoor church images.  Interestingly, some architectural styles are dropped even in cases where large-scale geometry is preserved.  The same set of unrelated (non-church) images as shown in \reffig{bedroom-ex} are shown.  When using the church model, the indoor reconstructions exhibit lower quality and are rendered to resemble outdoor scenes; the reconstructions of outdoor images recover more details.

\section{Discussion}
\noindent
We have proposed a way to measure and visualize mode-dropping in state-of-the-art generative models. 
\gissfull can compare the quality of different models and architectures, and provide insights into the semantic differences of their output spaces. Layer inversions allow us to further probe the range of the generators using natural photographs, revealing specific objects and styles that cannot be represented. By comparing labeled distributions with one another, and by comparing natural photos with imperfect reconstructions, we can identify specific objects, parts, and styles that a generator cannot produce.

The methods we propose here constitute a first step towards analyzing and understanding the latent space of a GAN and point to further questions.
Why does a GAN decide to ignore classes that are more frequent than others in the target distribution (e.g.~``person'' vs.~``fountain'' in \reffig{teaser})? How can we encourage a GAN to learn about a concept without skewing the training set? What is the impact of architectural choices?
Finding ways to exploit and address the mode-dropping phenomena identified by our methods are questions for future work.

\section*{Acknowledgements}
\vspace{-3pt}%
\noindent{\small We are grateful for the support of the MIT-IBM Watson AI Lab, the DARPA XAI program FA8750-18-C000, NSF 1524817 on Advancing Visual Recognition with Feature Visualizations, NSF BIGDATA 1447476, the Early Career Scheme (ECS) of Hong Kong (No.24206219) to BZ, and a hardware donation from NVIDIA.}


\newpage\clearpage

{\small
\bibliographystyle{ieee_fullname}
\bibliography{main.bib}
}

\end{document}


\title{Supplemental Materials for \\
Seeing What a GAN Cannot Generate}

\maketitle

\renewcommand{\thefigure}{S.\arabic{figure}}
\setcounter{figure}{0}
\renewcommand{\thetable}{S.\arabic{table}}
\setcounter{table}{0}
\renewcommand{\thesection}{S.\arabic{section}}
\setcounter{section}{0}

\begin{figure*}[!b]
\begin{minipage}{0.499\textwidth}
\centering
\textsf{\hspace{0.4in} \scriptsize LSUN Bedroom Dataset Sampled Twice}
\includegraphics[width=\columnwidth]{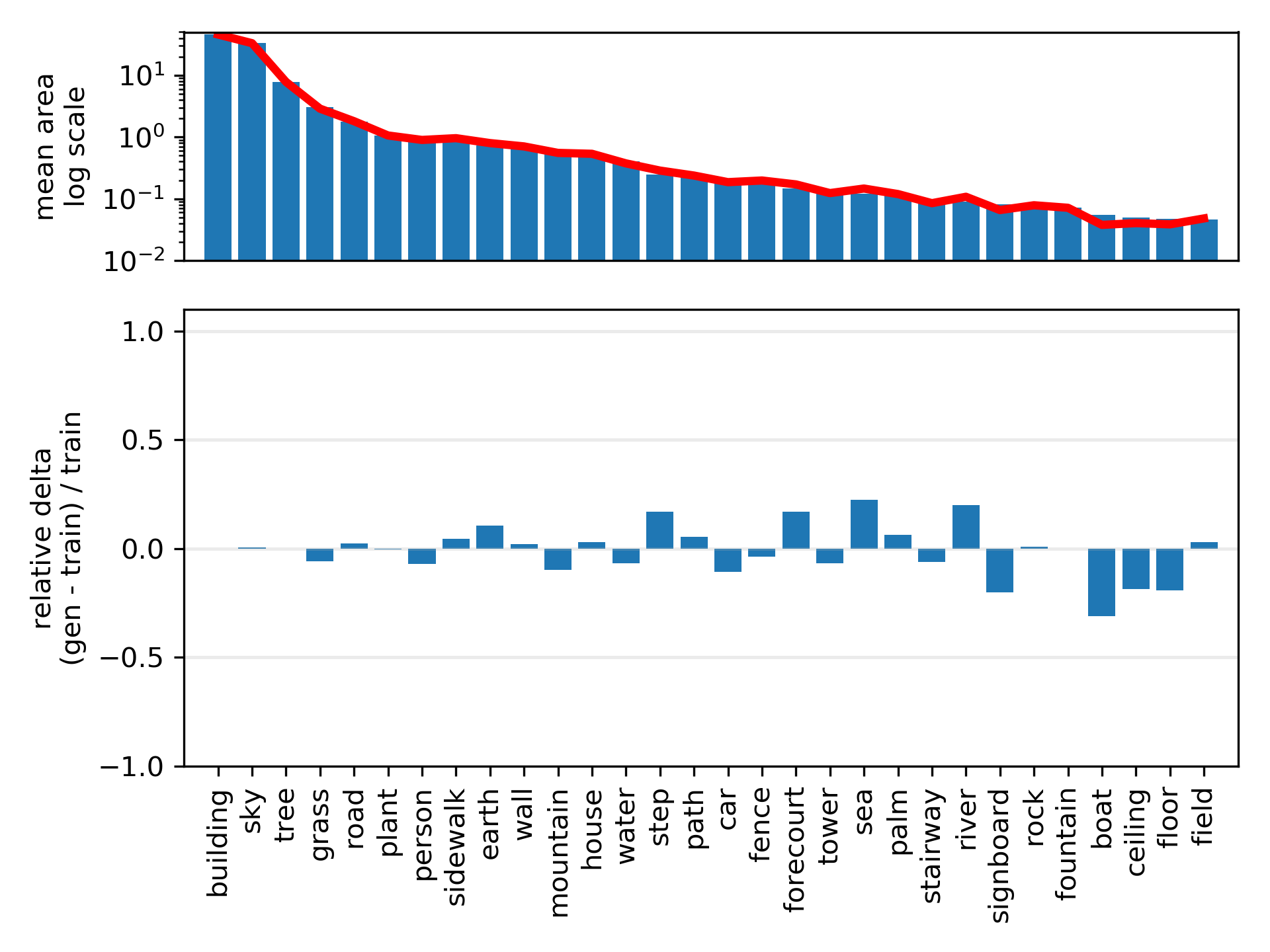}%
\end{minipage}%
\begin{minipage}{0.499\textwidth}
\centering
\textsf{\hspace{0.4in} \scriptsize LSUN Church Dataset Sampled Twice}
\includegraphics[width=\columnwidth]{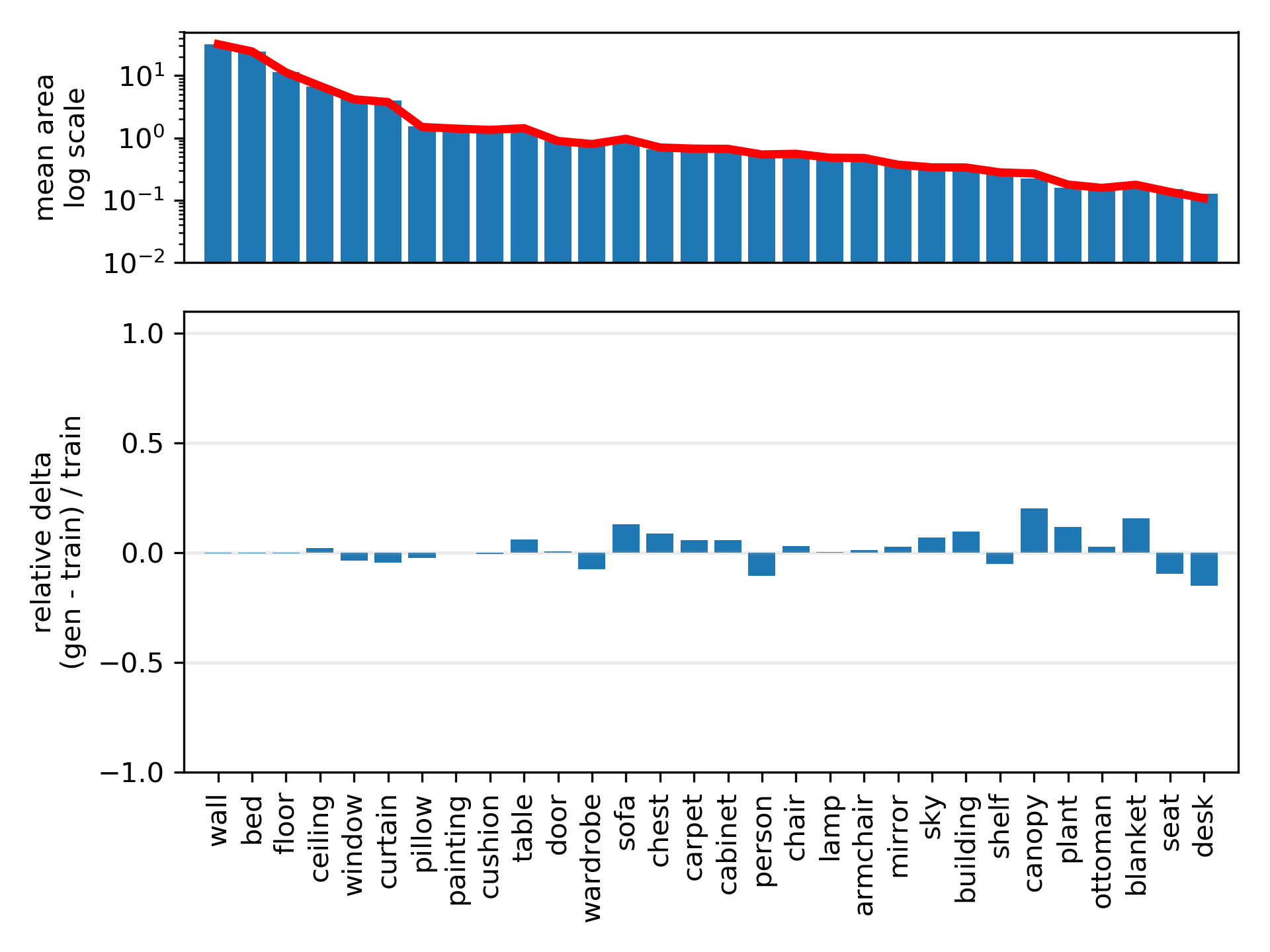}%
\end{minipage}
\caption{Sensitivity test for \gissfull.  At left, two subsamples of the LSUN outdoor church training set are compared.  At right, two subsamples of the LSUN bedroom training set are compared.  Each chart compares two different random samples of $10,000$ images from the same dataset.  An infinite-sized sample would show no differences; the observed differences reveal small measurement noise introduced by the finite sampling process.}
\lblfig{zero-fsd}
\end{figure*}
\section{Supplemental Materials}

\subsection{Sensitivity measure}
\gissfull are computed using sample statistics, so the estimated statistics will vary when the data is resampled.  Sampling error can be reduced by using a larger number of samples.  To estimate the sampling error of our measurements at the $10{,}000$ sample size used in our paper, \reffig{zero-fsd} and Table~\ref{tab:zero-fsd} use histograms and \fssshort to measure the difference between two different samples of the same data set.  Measurements are done for the LSUN outdoor church and the LSUN bedroom data sets~\cite{yu2015lsun}.
\begin{table}[H]
   \caption{Measured Sensitivity in \fssfull.} 
   \label{tab:zero-fsd}
   \small 
   \centering 
   \begin{tabular}{lcr} 
   \toprule[\heavyrulewidth]\toprule[\heavyrulewidth]
   \textbf{Data set} & \textbf{\fssshort vs self} \\ 
   \midrule
   LSUN outdoor church & 2.57 \\
   LSUN bedrooms & 5.57 \\
   \bottomrule[\heavyrulewidth] 
   \end{tabular}
   \vspace{-0.1in}
\end{table}

\newpage
\begin{figure*}[t!]
\centering
\includegraphics[width=\textwidth]{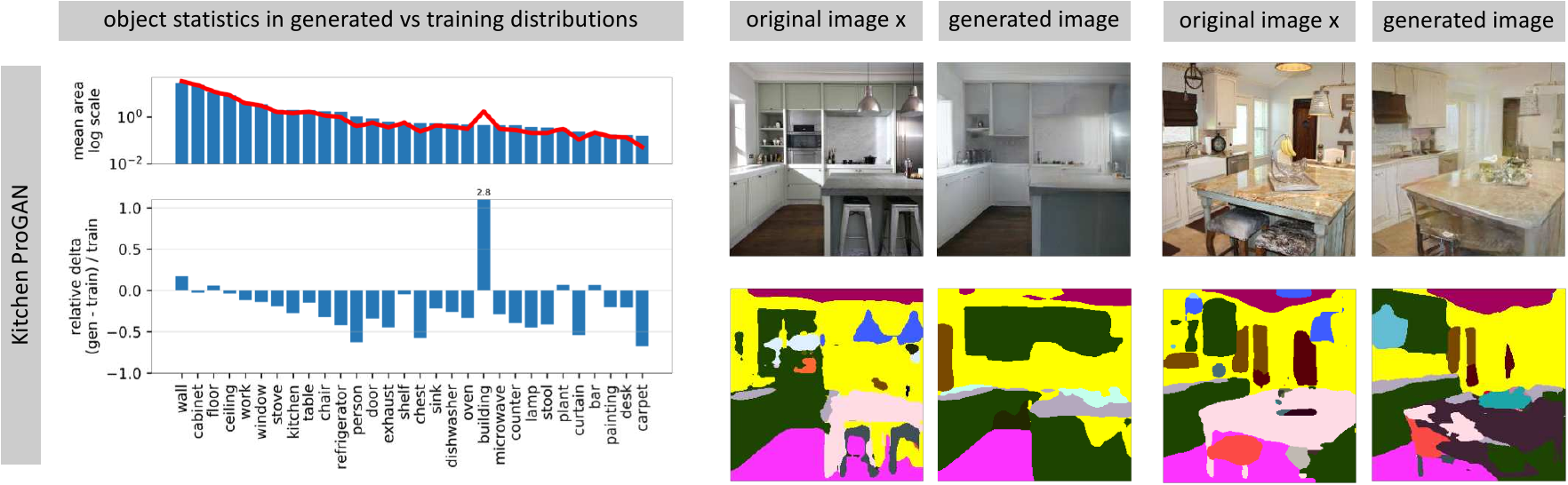}
\caption{Analysis of the differences in semantic object distribution between target and generated images for a Progressive GAN trained on LSUN kitchens. Chairs, stove exhausts, and other objects are underrepresented.}\lblfig{kitchen-analysis}
\end{figure*}
\begin{figure*}[t!]
\centering
\includegraphics[width=\textwidth]{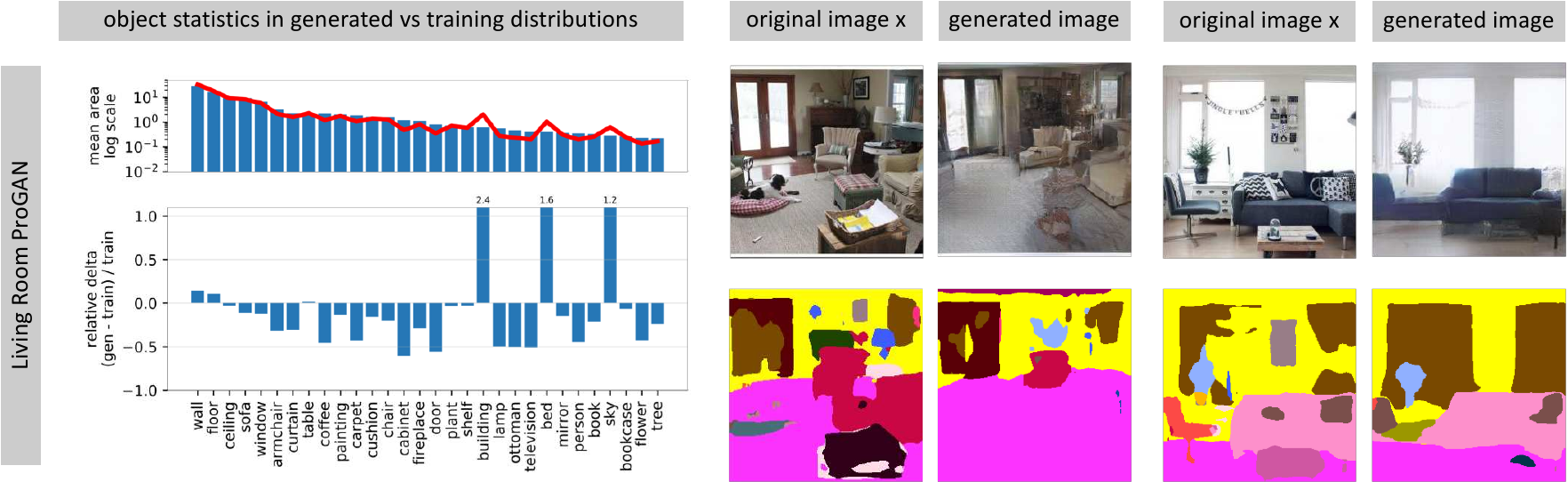}
\caption{Analysis of the differences in semantic object distribution between target and generated images for a Progressive GAN trained on LSUN living rooms. Some categories of furniture such as coffee tables and ottomans are omitted.}\lblfig{livingroom-analysis}
\end{figure*}
\begin{figure*}[t!]
\centering
\includegraphics[width=\textwidth]{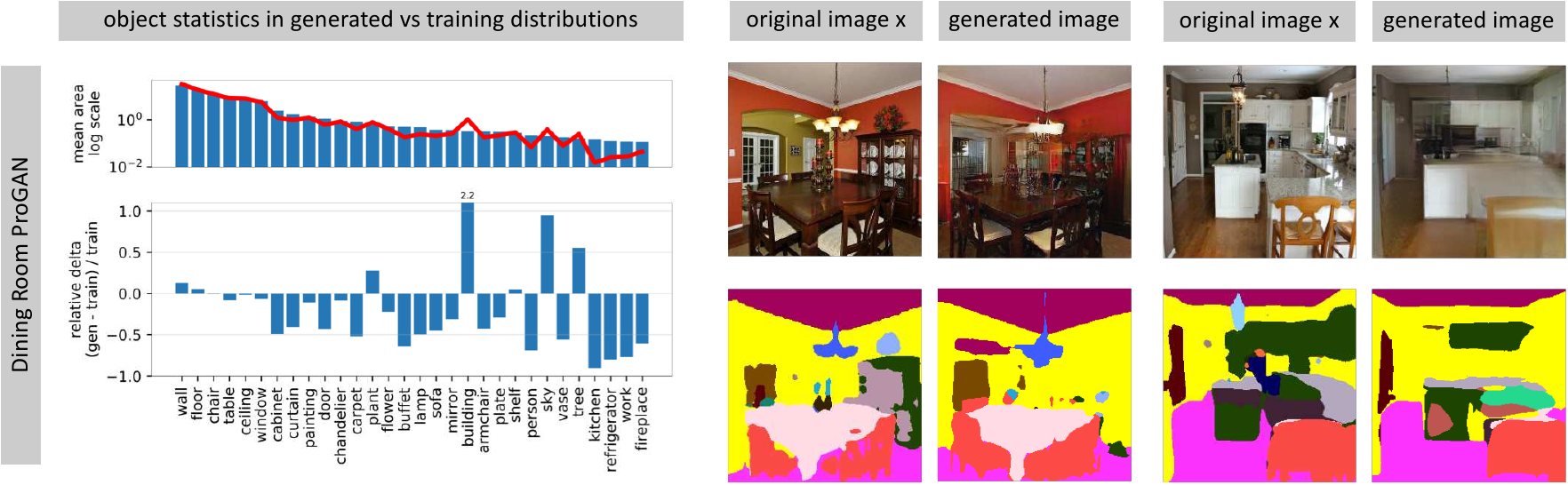}
\caption{Analysis of the differences in semantic object distribution between real and generated images for a Progressive GAN trained on LSUN dining rooms.  Dining rooms that include kitchens have lost many details.}\lblfig{diningroom-analysis}
\end{figure*}
\subsection{Analysis of unseen classes for additional GAN models}
Here we present examples of analysis of differences between generated and target semantic classes for several Progressive GAN models.  \reffig{kitchen-analysis} shows a Progressive GAN model trained on kitchens; \reffig{livingroom-analysis} shows a model for living rooms, \reffig{diningroom-analysis} shows a model for dining rooms. 

\subsection{Additional qualitative results on inversion}
Figure 5 in the main paper compares inversion methods quantitatively and includes only one image of each type (generated and photograph) for qualitatively comparing our inversion method with baselines and ablations.  In this section we present a larger number of images comparing the methods using reconstructions of church and bedroom models.  \reffig{church-ablations} shows reconstructions for several generated images as well as images from the validation set for LSUN.  \reffig{bedroom-ablations} shows the same for a bedroom model.
\begin{figure*}
\centering
\vspace{-9pt}%
\includegraphics[width=\textwidth]{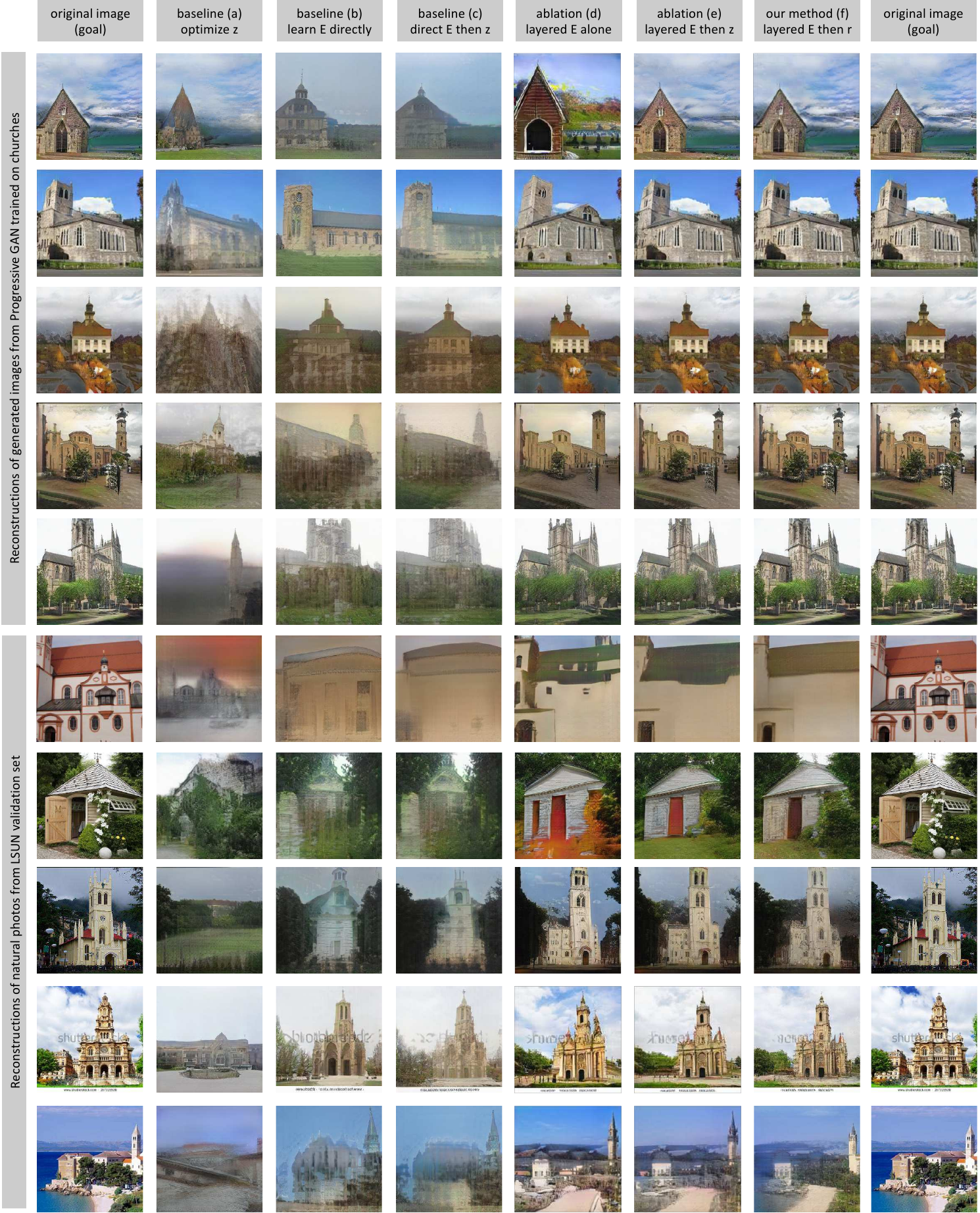}
\caption{Examples of reconstructions of both GAN-generated and holdout images, using several inversion methods.  The columns are the same as in Figure 5 in the main paper. (The target image is repeated to aid comparisons.)  At top are GAN-generated images which can be reconstructed nearly perfectly.  Below are natural photographs.  Dropped details reveal objects and styles that cannot be rendered by the GAN.}\lblfig{church-ablations}
\end{figure*}
\begin{figure*}
\centering
\vspace{-9pt}%
\includegraphics[width=\textwidth]{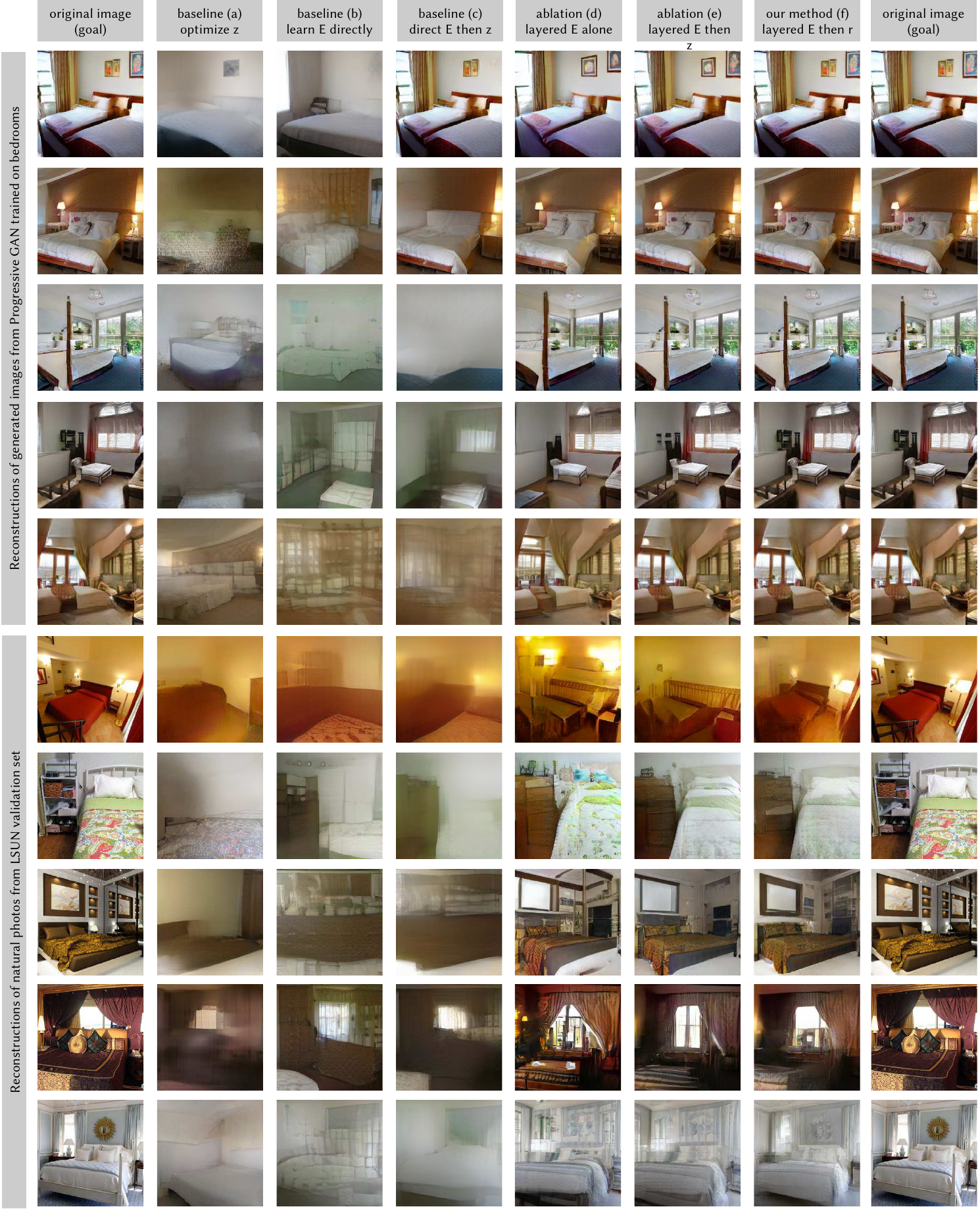}
\caption{Examples of reconstructions of both GAN-generated and holdout images, using several inversion methods.  The columns are the same as in Figure 5 in the main paper. (The target image is repeated to aid comparisons.)  At top are GAN-generated images, which can be reconstructed nearly perfectly.  Below are natural photographs.  Dropped details reveal objects and styles that cannot be rendered by the GAN.}
\lblfig{bedroom-ablations}
\end{figure*}